\documentclass[10pt,twocolumn,letterpaper]{article}

\usepackage{wacv}
\usepackage{times}
\usepackage{epsfig}
\usepackage{graphicx}
\usepackage{amsmath}
\usepackage{amssymb}

\usepackage{booktabs}
\usepackage{multirow}
\usepackage{adjustbox}
\usepackage{amssymb}
\usepackage{pifont}
\newcommand{\cmark}{\ding{51}}%
\newcommand{\xmark}{\ding{55}}%
\usepackage{amsmath,amssymb}

\DeclareMathOperator{\EX}{\mathbb{E}}

\def\wacvPaperID{1114} 

\wacvfinalcopy 

\ifwacvfinal
\def\assignedStartPage{1} 
\fi

\ifwacvfinal
\usepackage[breaklinks=true,bookmarks=false]{hyperref}
\else
\usepackage[pagebackref=true,breaklinks=true,colorlinks,bookmarks=false]{hyperref}
\fi

\ifwacvfinal
\else
\fi

\begin{document}

\title{iPerceive: Applying Common-Sense Reasoning to Multi-Modal Dense Video Captioning and Video Question Answering}

\author{Aman Chadha\\
Stanford University\\
{\tt\small hi@aman.ai}
\and
Gurneet Arora\\
University of Waterloo\\
{\tt\small gkarora@uwaterloo.ca}
\and
Navpreet Kaloty\\
Harvard University\\
{\tt\small nak764@g.harvard.edu}
}

\maketitle

\begin{abstract}
Most prior art in visual understanding relies solely on analyzing the ``what'' (e.g., event recognition) and ``where'' (e.g., event localization), which in some cases, fails to describe correct contextual relationships between events or leads to incorrect underlying visual attention. Part of what defines us as human and fundamentally different from machines is our instinct to seek causality behind any association, say an event $Y$ that happened as a direct result of event $X$. To this end, we propose iPerceive, a framework capable of understanding the ``why'' between events in a video by building a common-sense knowledge base using contextual cues to infer causal relationships between objects in the video. We demonstrate the effectiveness of our technique using the dense video captioning (DVC) and video question answering (VideoQA) tasks. Furthermore, while most prior work in DVC and VideoQA relies solely on visual information, other modalities such as audio and speech are vital for a human observer's perception of an environment. We formulate DVC and VideoQA tasks as machine translation problems that utilize multiple modalities. 
   By evaluating the performance of iPerceive DVC and iPerceive VideoQA on the ActivityNet Captions and TVQA datasets respectively, we show that our approach furthers the state-of-the-art. Code and samples are available at: 
  \url{iperceive.amanchadha.com}
\end{abstract}


\begin{figure}[t]
    \vspace {-4mm}
    \centering
    \includegraphics[width=0.85\linewidth]{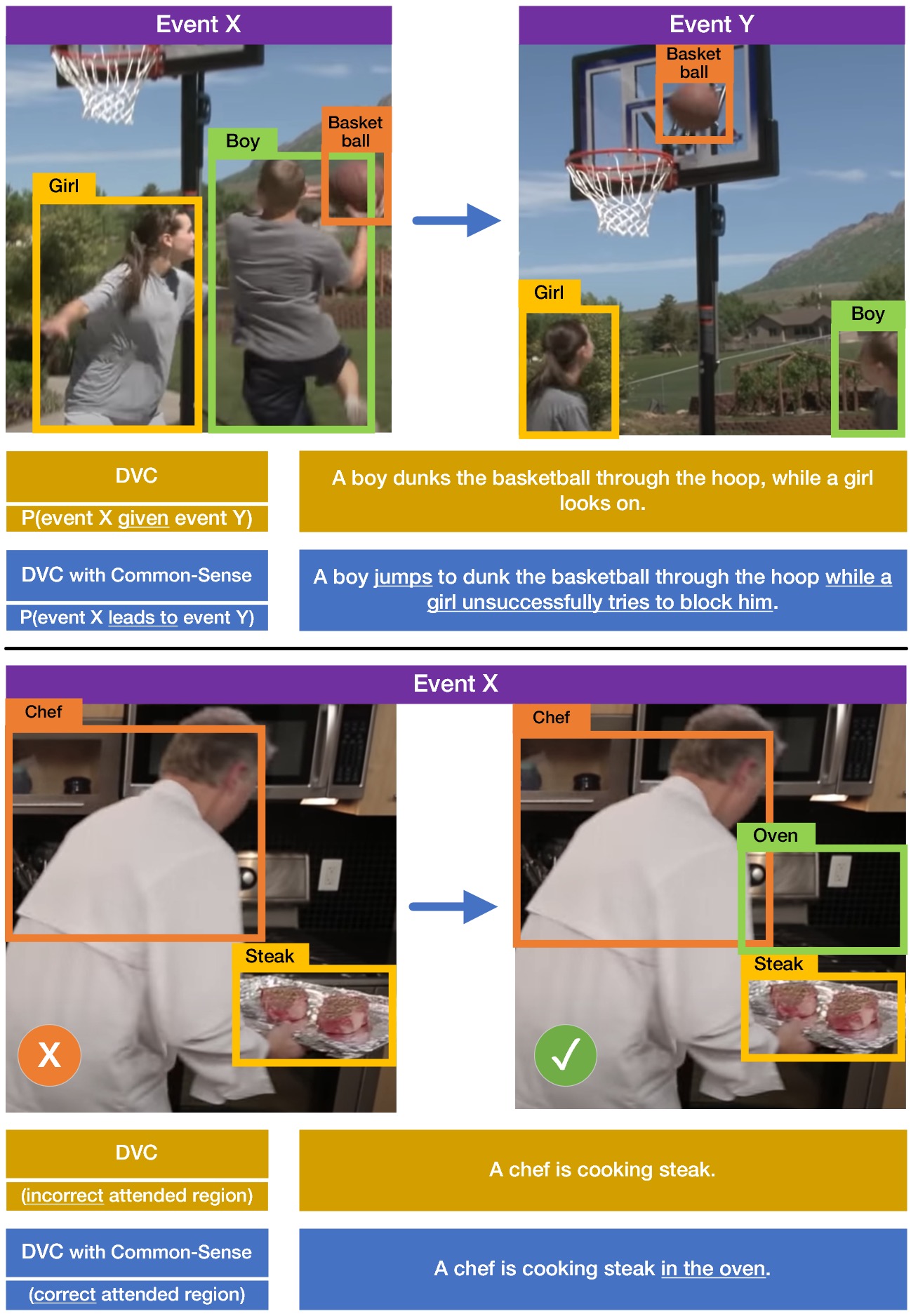} \vspace {1mm}
    \caption{Top: An example of a cognitive error in DVC. While the girl tries to block the boy's dunking attempt, him \emph{jumping} (event $X$) eventually \emph{leads} to him dunking the basketball through the hoop (event $Y$). Bottom: An example of incorrect attention 
    where conventional DVC approaches correlate a chef and steak to the activity of cooking \emph{without even attending} to the nearby oven. We used \cite{iashin2020multi} as our DVC baseline as it is the current state-of-the-art.}
    \label{causal}
\end{figure}

\section{Introduction}

Today’s computer vision systems are good at telling us the ``what'' (e.g., classification \cite{karpathy2014large}, segmentation \cite{grundmann2010efficient}) and “where” (e.g., detection \cite{li2003foreground}, localization \cite{tian2018audio}, tracking \cite{zhang2017deep}). Common-sense reasoning \cite{pearl2016causal}, which leads to the interesting question of ``why'', is a thinking gap in today's pattern learning-based systems which rely on the likelihood of observing object $Y$ given object $X$, $P(Y \vert X)$. 

Failing to factor in causality leads to the incorrect conclusion that the co-existence of objects $X$ and $Y$ might be attributed to spurious observational bias \cite{hendricks2018women, manjunatha2019explicit}. For e.g., if a \emph{keyboard} and \emph{mouse} are often observed on a table, the model learns to develop an ``association'' between the two. The underlying common-sense that the \emph{keyboard} and \emph{mouse} are parts of a computer would not be inferred, and in fact the duo would be wrongly associated as being part of a table. In the event that a \emph{keyboard} and \emph{mouse} are observed outside of a tabular setting, the model can commit a cognitive error. 
Prior work \cite{hendricks2018women, manjunatha2019explicit} has unraveled the spurious observational bias that models fall prey to. To alleviate this, we propose iPerceive -- a framework that utilizes contextual cues to establish a common-sense knowledge base from one of the most common ways humans acquire information: videos. 

Given the prevalence of visual information, video understanding is particularly important.
To that end, the task of dense video captioning (DVC) \cite{krishna2017dense} aims to temporally localize events from an untrimmed video and describe them using natural language. 
On the other hand, video question answering (VideoQA) is another challenging task in computer vision which requires significant expressive power from the model to distill visual events and their relations using linguistic concepts.


The vast majority of research in the field of DVC \cite{krishna2017dense, zhou2018end, wang2018bidirectional} and VideoQA \cite{zeng2017leveraging, hu2017learning, antol2015vqa, singh2019towards, anderson2018bottom, lu2019vilbert, lu2016hierarchical} generates captions purely based on visual information. However, given the fact that auditory feedback is an essential aspect of human communication, unsurprisingly, almost all videos include an audio track and some also include a speech track, both of which could provide vital cues for understanding the context of the event. 
Inspired by Iashin et al. \cite{iashin2020multi}, our DVC model consumes the video, audio and speech modality for the caption generation process. Similarly, our VideoQA implementation utilizes video and text (in the form of dense captions, subtitles and QA).

The task of DVC can be decomposed into two parts: event detection and event description. Existing methods tackle this using a module for each of these sub-tasks, and either train the two modules independently \cite{iashin2020multi} or in alternation \cite{krishna2017dense, wang2018bidirectional, chen2018ruc+}. This restricts model ``wiggle'', i.e., since events in a video sequence and the generated language are closely related, the language information should ideally be able to help localize events in the video. 
We address this by performing end-to-end training of our DVC model. While \cite{iashin2020multi} present a detailed study of the merits of using multiple modalities for DVC, they do not implement an end-to-end trainable system and train the captioning module on ground-truth event proposals. To this end, we utilize an end-to-end trainable model similar to \cite{zhou2018end} -- this fosters consistency between the content in the proposed video segment and the semantic information in the language description. Note that Zhou et al. \cite{zhou2018end} rely solely on the visual modality and hence, do not implement multi-modal DVC. Our approach blends multi-modal DVC with end-to-end learning.


We present iPerceive, a framework that generates common-sense features by inferring the causal relationships between events in videos using contextual losses as self-supervised training mechanisms. This enables the model to seek intrinsic causal relationships between objects within events in a video sequence. Furthermore, we offer hands-on evaluation of iPerceive using the tasks of DVC and VideoQA as case-studies. iPerceive DVC is a system that utilizes common-sense features and offers an end-to-end trainable multi-modal architecture that enables coherent dense video captions. Next, we propose an enhanced multi-modal architecture called iPerceive VideoQA that utilizes common-sense feature generation using iPerceive and dense captions using iPerceive DVC as its building blocks.

Our key contributions are centered around common-sense reasoning for videos, which we envision as a step towards human-level causal learning. 
Wang et al. \cite{wang2020visual} tackle the issue of observational bias in the context of images using common-sense generation, but applying a similar set of ideas to videos comes with its own set of challenges distinct from the image case. One observation is that events in videos can range across multiple time scales and can even overlap. Also, events can have causal relationships between themselves 
 that humans subconsciously perceive without any visible acknowledgment/feedback.
Humans naturally learn common sense in an unsupervised fashion by exploring the physical world, and until machines imitate this learning path, there will be a ``gap'' between man and machine. This requires us to build a knowledge base and acquire contextual information from temporal events in a video sequence to determine inherent causal relationships. These ``context-aware'' features can improve both the accuracy of contextual relationships as well as steer attention to the appropriate entities. Furthermore, videos are generally challenging to process compared to images owing to the sheer amount of data they contain. 

We employ several techniques to tackle the aforementioned nuances specific to videos (cf. Section \ref{idea-csvideos} for details). Furthermore, we offer a two-pronged evaluation of our proposed model by applying it to the challenging domains of DVC and VideoQA. A noteworthy point is that since our common-sense features can be generated in a self-supervised manner, they have a certain universality and are not limited to the realizations of DVC and VideoQA discussed in this work. As such, they can be easily adapted for other video-based vision tasks such as scene understanding \cite{hu2020probabilistic}, panoptic segmentation \cite{kim2020video}, etc.

\section{Related Work}

\subsection{Visual Common-Sense}
Current research in the field of building a common-sense knowledge base mainly falls into two categories: (i) learning from images \cite{yatskar2016stating, vedantam2015learning, zhu2014reasoning} and (ii) learning actions from videos \cite{goyal2017something}. While the former limits learning to human-annotated knowledge which restricts its effectiveness and outreach, the latter is essentially learning from correlation. 

\subsection{Causality in Vision}
There has been a recent surge of interest in coupling the complementary strengths of computer vision and causal reasoning \cite{pearl2016causal, pearl2014interpretation}. The union of these fields has been explored in several contexts, including image classification \cite{lopez2017discovering, chalupka2014visual}, reinforcement learning \cite{nair2019causal}, adversarial learning \cite{kocaoglu2017causalgan}, visual dialog \cite{qi2019two}, image captioning \cite{zhou2020more} and scene/knowledge graph generation \cite{tang2020unbiased, pan2020spatio}. While these methods offer limited task-specific causal inference, \cite{wang2020visual} offers a generic feature extractor for images.

\subsection{Video Captioning}
With the success of neural models in translation systems \cite{sutskever2014sequence}, similar methods became widely popular in video captioning \cite{yao2015describing, venugopalan2015sequence}. The core rationale behind this approach is to train two recurrent neural networks (RNNs) in an encoder-decoder fashion. Specifically, an encoder inputs a set of video features and accumulates its hidden state, which is passed on to a decoder for captioning.

\subsection{Dense Video Captioning}
A significant milestone in the domain of video understanding was reached when Krishna et al. \cite{krishna2017dense}, inspired by the idea of the dense image captioning task \cite{johnson2015densecap}, introduced the problem of DVC and the idea of context-awareness to utilize both past and future context. They also released a new dataset called ActivityNet Captions which has propelled research in the field \cite{zhou2018end, wang2018bidirectional, li2018jointly}. Furthermore, \cite{wang2018bidirectional} proposed attentive fusion to differentiate captions from highly overlapped events.
With the recent success of Transformers \cite{vaswani2017attention} in the machine translation task, it was inevitable for them to enter the similarly-complex task of video understanding. Zhou et al. \cite{zhou2018end} adopted Transformers for DVC to alleviate the limitations of RNNs when modeling long-term dependencies in videos. 

\subsection{Multi-Modal Dense Video Captioning}
Several attempts have been made to incorporate additional cues like audio and speech \cite{rahman2019watch, hessel2019case} for the dense video captioning task. Rahman et al. \cite{rahman2019watch} utilized the idea of cycle-consistency to build a model with visual and audio inputs. 
Hessel et al. \cite{hessel2019case} and Shi et al. \cite{shi2019dense} employ a Transformer architecture to encode both video frames and speech segments to generate captions for instructional cooking videos where speech and the captions are usually well-aligned with the visual content \cite{miech2019howto100m}. While they achieve stellar results, their model fails to generalize to real-world videos where speech and captions can have a ``gap'' with the visual inputs. Iashin et al. \cite{iashin2020multi} tackle the problem of multi-modal DVC using a Transformer-based architecture and renders great results, but do not incorporate the concept of end-to-end training introduced in \cite{zhou2018end}.

\subsection{Visual/Video Question Answering}
The allied tasks of Visual Question Answering (VQA) and VideoQA involve the important ability of understanding visual information conditioned on language. While QA based on a single image, i.e., VQA, has been well explored \cite{hu2017learning, antol2015vqa, singh2019towards, anderson2018bottom, lu2019vilbert, lu2016hierarchical}, the field of VideoQA is now picking up intense momentum \cite{zeng2017leveraging, kim2020dense, fan2019heterogeneous, lei2019tvqa+, le2020hierarchical}. While VQA relies on spatial information, VideoQA requires an understanding of both spatial and temporal information, making the task of VideoQA inherently much more complex compared to VQA. Within VideoQA, Zeng et al. \cite{zeng2017leveraging} have explored using non-dense image captions. However, there exists limited research that utilizes dense captions to help improve the temporal localization of videos.
Kim et al. \cite{kim2020dense} tackle the task of VideoQA using DIC, while we take it a step further by utilizing DVC with common-sense features.

\begin{figure*}[!ht]
    \vspace {-4mm}
    \centering
    \includegraphics[width=0.85\linewidth]{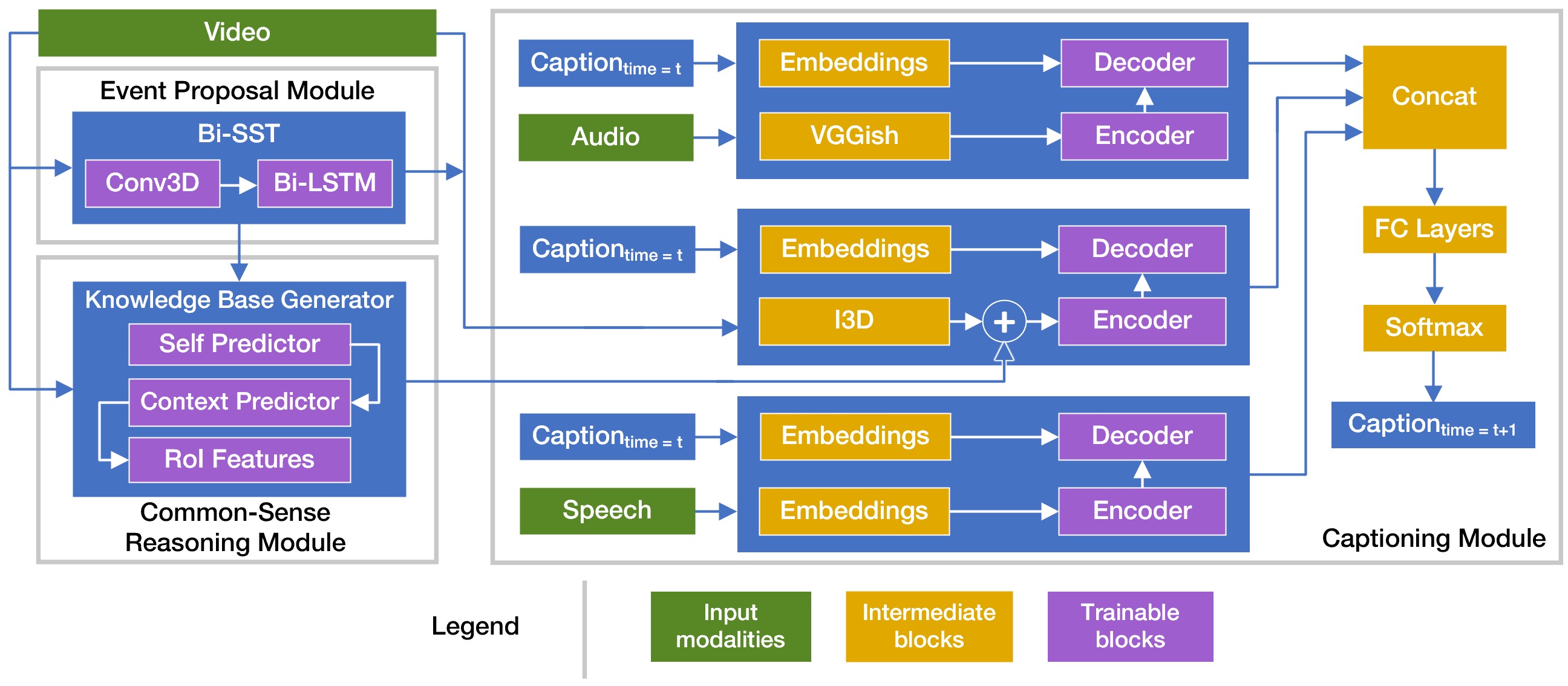} \vspace {1mm}
    \caption{Architectural overview of iPerceive DVC. 
    iPerceive DVC generates common-sense vectors from the temporal events that the event proposal module localizes (left). Features from all modalities are sent to the corresponding encoder-decoder Transformers (middle). Upon fusing the processed features we finally output the next word in the caption using the distribution over the vocabulary (right).}
    \label{dvcarch}
\end{figure*}

\section{iPerceive DVC: Proposed Framework}

\subsection{Top Level View}

Fig. \ref{dvcarch} outlines the goals of iPerceive DVC: (i) temporally localize a set of events in a video, (ii) build a knowledge base for common-sense reasoning and, (iii) produce a textual description using audio, visual, and speech cues for each event. To this end, we apply a three-stage approach. In this paper, we limit our discussion of implementational details to the blocks that we adapt to enable the common-sense reasoning aspect of iPerceive DVC. For further details on the building blocks of our architecture, we refer the curious reader to our baseline for the DVC task \cite{iashin2020multi}.

\subsection{Event Proposal Module}

We localize the temporal events in the video 
using the bidirectional single-stream (Bi-SST) network \cite{wang2018bidirectional}. Bi-SST applies 3D convolutions (C3D) \cite{tran2015learning} to video frames and passes on the extracted features to a Bi-directional LSTM \cite{hochreiter1997long} network. The LSTM accumulates visual cues from past and future context over time and predicts the endpoints of each event in the video along with its confidence scores. 
The output of the LSTM feeds the common-sense reasoning module to convey the set of events in the input video to extract common-sense features for.

\subsection{Common-Sense Reasoning: Videos vs. Images}\label{idea-csvideos}

\subsubsection{Computational Complexity}

It is common knowledge that compared to images, tasks involving videos need an exponentially larger set of resources, both in terms of compute and time. To make the task of common-sense feature generation for videos computationally tractable, we only generate features for a frame when we detect a change in the environmental ``setting'' going from one frame to the next, within a particular localized event. Specifically, we check for changes in the set of object labels in a scene and only generate common-sense features if a change is detected; if not, we re-use the common-sense features from the last frame.

\subsubsection{On-the-fly RoI Generation}

Note that the architecture proposed in \cite{wang2020visual} essentially serves as an improved visual region encoder given region(s) of interest (RoIs) in an image. As such, it assumes that an RoI exists and is available at train time. This greatly limits its usability to models that inherently extract RoIs (such as Mask R-CNN \cite{he2017mask}, Faster R-CNN \cite{ren2015faster}, etc.), and thus reduces its effectiveness with use-cases beyond object detection, such as DVC and VideoQA. As such, we extend their work by utilizing a pre-trained Mask R-CNN \cite{he2017mask} model to generate RoIs for frames within each event that has been localized by the event proposal module before passing them onto the common-sense reasoning module.

\subsection{Common-Sense Reasoning Module}\label{idea-causal}


Common-sense reasoning via self-supervised representation learning serves as an improved visual region encoder for subsequent processing. In the context of developing a common-sense knowledge base, one of the biggest challenges involved is determining causal context: how do you figure out the cause and effect relationship between objects and thus re-align the model's context? 
Common-sense reasoning is based on causality which relies on $P(Y \vert do(X))$ \cite{pearl2016causal}. This is fundamentally different from what prior work in the domain of DVC or VideoQA, or video understanding in general, relies on: the conventional likelihood, $P(Y \vert X)$.

Building upon the approach in \cite{wang2020visual}, we carry out the following deliberate ``borrow-put'' experiment for a given frame within an event localized by the proposal module: (1) ``borrow'' non-local context, say an object $Z$ from another event, (2) ``put'' $Z$ in the context of object $X$ and object $Y$ and, (3) test if object $X$ still causes the existence of object $Y$ given $Z$. 
This experiment helps determine if the chance of $Z$ is independent on $X$ or $Y$. 
Thus, by using $P(Y \vert do(X))$ as the learning objective instead of $P(Y \vert X)$, the observational bias from the ``apparent'' context can be alleviated. As such, the training objective of the common-sense reasoning module is the proxy task of predicting the contextual objects of an event.

Our visual world contains several confounding agents $z$ $\in$ $Z$ that add spurious observational bias around objects $X$ and $Y$ and hinder common-sense development. This limits the model's learning using the traditional likelihood $P(Y \vert X)$, which can be 
defined \cite{wang2020visual} using Bayes' rule as:
\begin{equation}
P(Y|X) = \sum\limits_z {P(Y|X,z)P(z|X)}
\end{equation}
where, the confounder $z$ introduces spurious bias via $P(z \vert X)$.

Since we can hardly identify all confounders in the real world, we approximate the set of confounder objects $Z$ to a fixed confounder dictionary in the shape of a $N \times d$ matrix for practical use, where $N$ is the number of classes in the dataset (e.g., 80 in MS-COCO \cite{lin2014microsoft}) and $d$ is the feature dimension of each RoI. Each entry $z \in Z$ is the averaged RoI feature, obtained using Faster R-CNN \cite{ren2015faster} for the samples in the dataset that belong to same class as $z$.

We similarly define the ``do'' operation by disrupting the causal link between $z$ and $X$ (and thus de-biasing $X$) as,
\begin{equation} \label{eq:1}
P(Y|do\left(X \right)) = \sum\limits_z {P(Y|X,z)P(z)}
\end{equation}

Each RoI $X$ is then fed into two sibling branches: (i) a self-predictor to predict the class of the ``center'' object $x \in X$ and, (ii) a context-predictor to predict the ``center'' object's context labels, $y_i \in Y$. The self-predictor outputs a probability distribution $p$ over $N$ categories. On the other hand, the context-predictor outputs a probability distribution for a pair of RoIs ($x$, $y_i$). The last layer of the network uses a Softmax classifier for label prediction: 
\begin{equation}
P(Y|do\left(X \right)) = \EX_{z}(Softmax(f(x,z)))
\end{equation}
where, $f(\cdot)$ calculates the logits for $N$ categories and $\EX_{z}$ is obtained by sampling $z$ over the set of confounders $Z$.

We utilize the normalized weighted geometric mean (NWGM) to approximate the above expectation. A detailed discourse for NWGM has been provided in \cite{wang2020visual}. Furthermore, for an in-depth visual treatment on how the features learned by correlation $P(Y \vert X)$ and causal intervention $P(Y \vert do(X))$ differ, along with examples on how this leads to building up of common-sense, we direct the avid reader to \cite{wang2020visual}. These common-sense features are then paired with the corresponding visual features for each localized event and sent downstream for captioning.

\subsection{Captioning Module}\label{idea-caption}

Given an event proposal and its common-sense vectors, the captioning module generates a caption using audio, visual and speech modalities. We formulate the captioning task as a machine translation problem and adapt the Transformer-based architecture from \cite{iashin2020multi}.
We use inflated 3D convolutions (I3D) \cite{carreira2017quo} to process visual modalities and the VGGish network \cite{hershey2017cnn} for audio modalities. We deploy an automatic speech recognition (ASR) system \cite{youtubecaps} to extract temporally-aligned speech transcriptions. These are juxtaposed alongside the video frames and the corresponding audio track, and fed in as input to our model. 
Features from each modality are then fed to individual Transformer models along with the generated caption so far. The output of each Transformer is fused and a probability distribution is obtained over the vocabulary to sample the next word until a special end token is obtained.

A notable point is that the self-attentive operation inherently has an indiscriminate correlation against our new learning objective $P(Y \vert do(X))$ \cite{wang2020visual}. Put differently, self-attention implicitly applies conventional likelihood $P(Y \vert X)$, which contradicts causal reasoning $P(Y \vert do(X))$. Furthermore, given that the computation of self-attention is expensive, especially in the case of multiple heads, early concatenation of common-sense features significantly slows down training. We thus omit the self-multi-headed attention component in the encoder of \cite{iashin2020multi}.


\begin{figure*}[ht]
    \vspace {-4mm}
    \centering
    \includegraphics[width=0.85\linewidth]{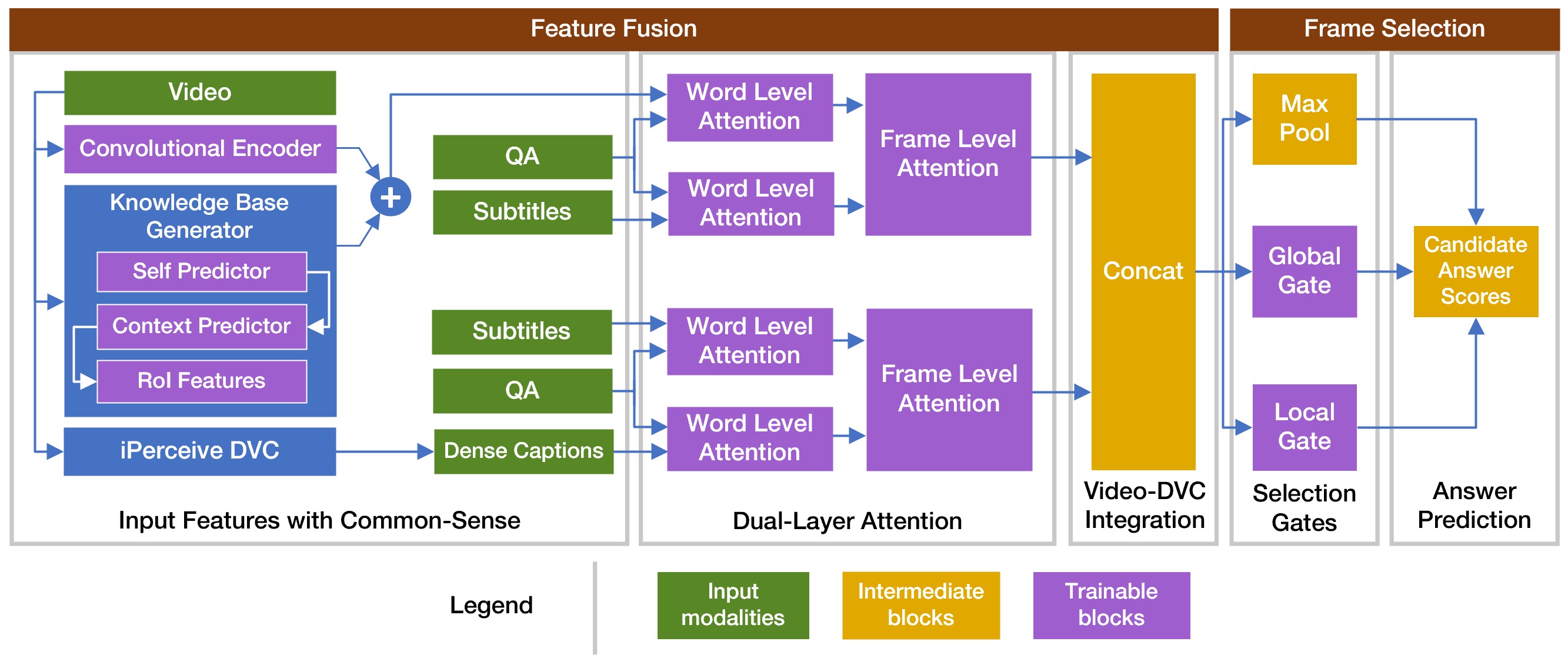} \vspace {1mm}
    \caption{Architectural overview of iPerceive VideoQA. 
    Our model consists of two main components: feature fusion and frame selection. For feature fusion, we encode features using a convolutional encoder, generate common-sense vectors from the input video sequence, and extract dense captions using iPerceive DVC (left). Features from all modalities (video, dense captions, QA and subtitles) are then fed to dual-layer attention: word/object and frame-level (middle). Upon fusing the attended features, we calculate frame-relevance scores (right).}
    \label{vidqaarch}
\end{figure*}

\subsection{Loss Functions}

iPerceive DVC uses a four-fold training loss: (i) 
cross entropy $MCE$ as proposal loss $L_{p}$ to balance positive and negative proposals, (ii) multi-task common-sense reasoning loss $L_{cs}$, (iii) binary cross entropy $BCE$ as mask prediction loss $L_{m}$ and, (iv) cross entropy $MCE$ across all words in every sentence as captioning loss $L_{c}$. 

\subsubsection{Proposal Loss}
\begin{equation}
{L_p} = MCE(c,t,X,y)
\end{equation}
where, $c$ is the prediction score at time $t$, $X$ is the input video and $y$ is the ground truth label with an acceptable intersection over union (IoU).

\subsubsection{Common-Sense Reasoning Losses} \label{csloss}
The common-sense reasoning losses are a set of auxiliary self-supervised losses that help capture knowledge about the co-occurence of objects within events in a video scene. 

Consider a video frame which consists of a ``center'' object and a ``context'' object. For a ``center'' object $x \in X$ in the video frame at time $t$, the self-predictor loss $L_{self}$ can be defined using negative log likelihood as,
\begin{equation}
{L_{self}}(p,{x^c},t) = -log(p[x^c])
\end{equation}
where, $p$ is the probability distribution output of the self-predictor over $N$ categories for $X$;
$x^c$ is the ground-truth class of RoI $X$. 

Similarly, for a ``context'' object $y_i \in Y$ in the video frame at time $t$, the context-predictor loss $L_{cxt}$ is defined for a pair of RoI feature vectors ($x$, $y_i$) using negative log likelihood as,
\begin{equation}
{{L_{cxt}}({p_i},y_i^c,t)} = -log(p_{i}[y_{i}^{c}])
\end{equation}
where, $y_i^c$ is the ground-truth label for $y_i$; $p_i$ is calculated using $p_i = P(Y_i|do(X))$ in Eq. \ref{eq:1} and $p_i = (p_i[1], ... , p_i[N])$ is the probability over $N$ categories. 

The overall multi-task loss for each RoI $X$ is,
\begin{equation}
{L_{cs}} = {L_{self}} + \frac{1}{K}\sum\limits_i {{L_{cxt}}}
\end{equation}

\subsubsection{Mask Prediction Loss}
\begin{equation}
{L_m} = BCE(Bin({S_a},{\rm{ }}{E_a},{\rm{ }}t),{\rm{ }}{f_M}({S_a},{\rm{ }}{E_p},{\rm{ }}{S_a},{\rm{ }}{E_a},{\rm{ }}t))
\end{equation}
where, 
$Bin(.)$ is `1' if $t$ $\in$ $[S_a, E_a]$ and `0' otherwise;
$f_M$ is a differentiable mask for time $t$ \cite{zhou2018end};
($S_p$, $E_p$) are the start and end times of the event;
($S_a$, $E_a$) are the start and end positions of the anchor.

\subsubsection{Captioning Loss}
\begin{equation}
{L_c} = MCE(w'_t, w_t)
\end{equation}
where $w'_t$ is the ground truth word at time $t$.

\subsubsection{Overall Loss Formulation}
The final loss $L$ is a combination of the individual losses, 
\begin{equation}
L = {\lambda _1}{L_p} + {\lambda _2}{L_{cs}} + {\lambda _3}{L_m} + {\lambda _4}{L_c}
\end{equation}
where $\lambda_{1-4}$ weigh the 
individual loss components.


\section{iPerceive VideoQA: Proposed Framework}

\subsection{Top Level View}

Building upon the architecture proposed by \cite{kim2020dense}, we propose iPerceive VideoQA, a model that uses common-sense knowledge to perform VideoQA. We utilize dense captions using iPerceive DVC to offer the model additional telemetry to correlate objects identified from video frames and their salient actions expressed through dense captions. 
Similar to iPerceive DVC, we limit our discussion of implementational details in this study to the blocks that we adapt to enable the common-sense reasoning aspect of iPerceive VideoQA. We refer the curious reader to our baseline for the VideoQA task \cite{kim2020dense} for details on the building blocks of our architecture.

Fig. \ref{vidqaarch} outlines the goals of iPerceive VideoQA: (i) build a common-sense knowledge base, (ii) extract features from multiple modalities: video and text (in the form of dense captions, subtitles and QA) and, (iii) implement the relevant-frames selection problem as a multi-label classification task. As such, we apply a two-stage approach.

\subsection{Feature Fusion}
\subsubsection{Feature Generation Module}

Leveraging the approach in \cite{kim2020dense}, we extract features from multiple modalities viz. video, dense captions, question-answer (QA) pairs and subtitles. We encode the visual features using a convolutional encoder.
To generate dense captions, we utilize iPerceive DVC and operate it at a frame-level to derive dense captions for the current frame. 
We create five hypotheses by concatenating a question feature with each of five answer features, and we pair each visual frame feature with temporally neighboring subtitles.

\subsubsection{Common-Sense Reasoning Module}

We utilize the common-sense generation module proposed in Section \ref{idea-causal} to generate common-sense vectors corresponding to each frame of the input video. 
iPerceive VideoQA 
concatenates common-sense features with the features extracted from the convolutional encoder and sends the output downstream.

\subsubsection{Dual-Layer Attention}

\textbf{Word/Object-Level Attention:} The visual features for each frame are combined with the textual features (QA and subtitles) using word/object-level attention, following the approach in \cite{kim2020dense}. Separately, we also combine the DVC features with the textual features in a similar manner. To this end, we calculate similarity matrices \cite{seo2016bidirectional} from (i) QA/subtitle and QA/visual features and, (ii) QA/subtitle and QA/DVC features, respectively. Attended subtitle features are obtained from the similarity matrices.

\textbf{Frame-Level Attention:} The fused features from word/object-level attention are integrated frame-wise via frame-level attention. Similar to the idea behind word/object-level attention, a similarity matrix is calculated which is used to compute the attended frame-level features.

\subsubsection{Video-DVC Integration Module}
\cite{kim2020dense} implements self-cross attention to amalgamate information from the dual-layer attended visual and dense caption features, both of which have been fused with QA and subtitles. As discussed in Section \ref{idea-causal}, due to the challenges associated with generating common-sense features using a model that implements self-attention, we carry out concatenation of video and dense caption features. 

\begin{figure*}[!ht]
    \centering
    \includegraphics[width=0.85\linewidth]{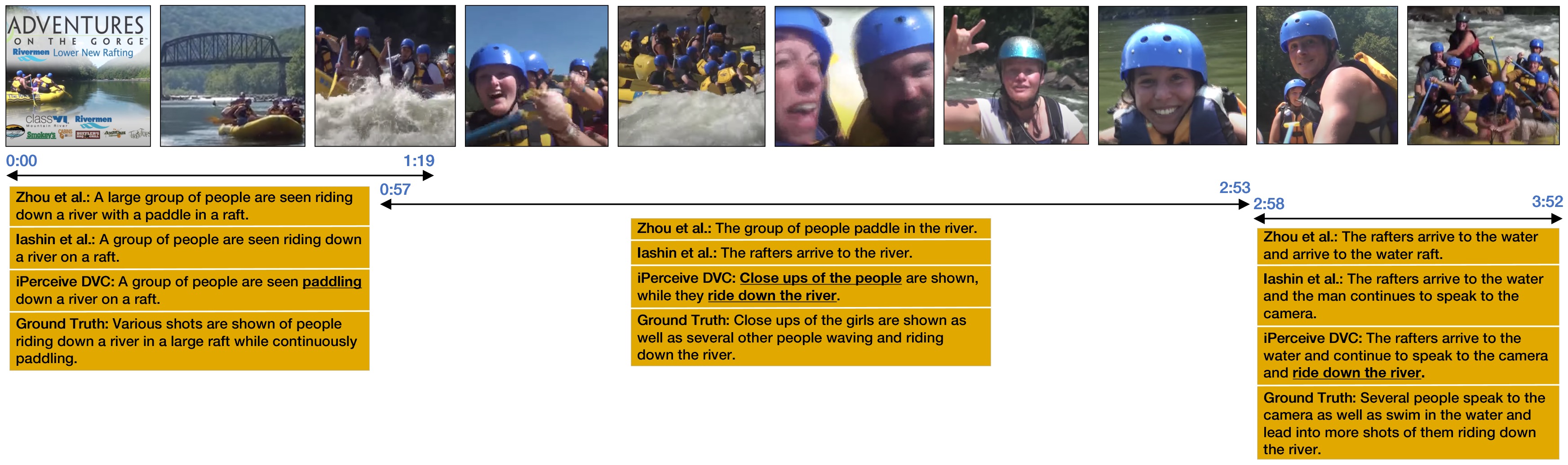} 
    \caption{Qualitative sampling of iPerceive DVC: Captioning results for a sample video from the ActivityNet Captions validation set.}
    \label{DVCsample}
\end{figure*}

\subsection{Frame Selection}
\subsubsection{Selection Gates}
Similar to \cite{kim2020dense}, we utilize gates to selectively control the flow of information and ensure only relevant information propagates through to the classifier. As such, we use a fully-connected layer to get frame-relevance scores that indicate the appropriateness quotient of each frame for answering a particular question. 
From the logits for the five candidate answers, we choose the highest value as our prediction.

\subsection{Loss Functions}

iPerceive VideoQA uses a four-fold training loss: (i) multi-task common-sense reasoning loss $L_{cs}$, (ii) softmax cross-entropy loss as answer selection loss $L_{ans}$, (iii) balanced binary cross entropy as frame-selection loss $L_{fs}$ and, (iv) in-and-out frame score margin $L_{io}$. We adopt similar multi-task common-sense reasoning losses ${L_{cs}}$ as in Section \ref{csloss}. The other loss components are similar to \cite{kim2020dense}.

\subsubsection{Overall Loss Formulation}
The final loss $L$ is a combination of the individual losses, 
\begin{equation}
L = {\lambda _1}{L_{cs}} + {\lambda _2}{L_{ans}} + {\lambda _3}{L_{fs}} + {\lambda _4}{L_{io}}
\end{equation}
where $\lambda_{1-4}$ weigh the 
individual loss components.

\section{Experimental Evaluation}
\subsection{iPerceive DVC}
\subsubsection{Dataset}
We train and assess iPerceive DVC using the ActivityNet Captions \cite{krishna2017dense} dataset,
using a train/val/test split of 50\%/25\%/25\%.
ActivityNet Captions contains 20k videos from YouTube. On an average, each video has 3.65 events that are 2 minutes long and are annotated by two different annotators using 13.65 words.
We report all results using the validation set (since no ground truth is available for the test set). 




\subsubsection{Metrics}
We use BLEU@N \cite{papineni2002bleu} and METEOR \cite{denkowski2014meteor}, which are common DVC evaluation metrics. 
We use the official evaluation script provided in \cite{krishna2017dense}.


\subsubsection{Comparison with Baseline Methods}
Tab. \ref{DVCresults} compares iPerceive DVC with the state-of-the-art. 
Algorithms were split into the ones which ``saw'' all training videos and others which trained on partially available data (since some YouTube videos which were part of the ActivityNet Captions dataset are no longer available). 
Fig. \ref{DVCsample} offers a qualitative comparison between iPerceive DVC and \cite{zhou2018end} and \cite{wang2020visual}, which were the best performing baselines for captioning and event localization, respectively. 

\begin{table}[!h]
\caption{Evaluation of iPerceive DVC on the ActivityNet Captions validation set using BLEU@N (B@N) and METEOR (M).}
\begin{adjustbox}{width=\columnwidth,center}
    \vspace {-2.5mm}
    \begin{tabular}
    {lcccccc}
        \toprule
\multirow{2}{*}{Method} & \multicolumn{3}{c}{GT Proposals} & \multicolumn{3}{c}{Learned Proposals} \\
& B@3 & B@4 & M & B@3 & B@4 & M \\\hline\hline
\multicolumn{7}{l}{\textbf{\textit{Seen full dataset}}} \\
Krishna et al. \cite{krishna2017dense} & 4.09 & 1.60 & 8.88 & 1.90 & 0.71 & 5.69 \\
Wang et al. \cite{wang2018bidirectional} & - & - & 10.89 & 2.55 & \textbf{1.31} & 5.86 \\
Zhou et al. \cite{zhou2018end} & 5.76 & 2.71 & 11.16 & 2.42 & 1.15 & 4.98 \\
Li et al. \cite{li2018jointly} & 4.55 & 1.62 & 10.33 & 2.27 & 0.73 & 6.93 \\
\hline\hline
\multicolumn{7}{l}{\textbf{\textit{Seen part of the dataset}}} \\
Rahman et al. \cite{rahman2019watch} & 3.04 & 1.46 & 7.23 & 1.85 & 0.90 & 4.93 \\
Iashin et al. \cite{iashin2020multi} & 4.12 & 1.81 & 10.09 & 2.31 & 0.92 & 6.80 \\
\textbf{iPerceive DVC} & 5.23 & 2.34 & 11.77 & 2.59 & 1.07 & 7.29 \\
Iashin et al. (all modalities) & 5.83 & 2.86 & 11.72 & 2.60 & 1.07 & 7.31 \\
\textbf{iPerceive DVC (all modalities)} & \textbf{6.13} & \textbf{2.98} & \textbf{12.27} & \textbf{2.93} & 1.29 & \textbf{7.87} \\
        \bottomrule
    \end{tabular}
\label{DVCresults}
\end{adjustbox}
\end{table}

\subsubsection{Ablation Analysis}
Tab. \ref{DVCab} shows ablation studies for iPerceive DVC to assess the impact of common-sense reasoning and end-to-end training as design decisions.

\begin{table}[!h]
\caption{Ablation analysis for iPerceive DVC.}
\begin{adjustbox}{width=0.8\columnwidth,center}
    \vspace {-2.5mm}
    \begin{tabular}
    {ccc}
        \toprule
Common-Sense Reasoning & End-to-End Training & METEOR\\
\hline\hline
\xmark & \xmark & 7.31\\
\xmark & \cmark & 7.42\\
\cmark & \xmark & 7.51\\
\cmark & \cmark & 7.87\\
        \bottomrule
    \end{tabular}
\label{DVCab}
\end{adjustbox}
\end{table}

\subsection{iPerceive VideoQA}
\subsubsection{Dataset}
We train and evaluate iPerceive VideoQA using the TVQA dataset \cite{lei2018tvqa} which consists of video frames, subtitles and QA pairs from six TV shows. The train/val/test splits for TVQA are 80\%/10\%/10\%. Each sample has five candidate answers with one as the ground-truth. TVQA is thus a classification task which can be evaluated based on accuracy.


\subsubsection{Comparison with Baseline Methods}
\begin{figure}[!hb]
    \centering
    \includegraphics[width=1.0\columnwidth]{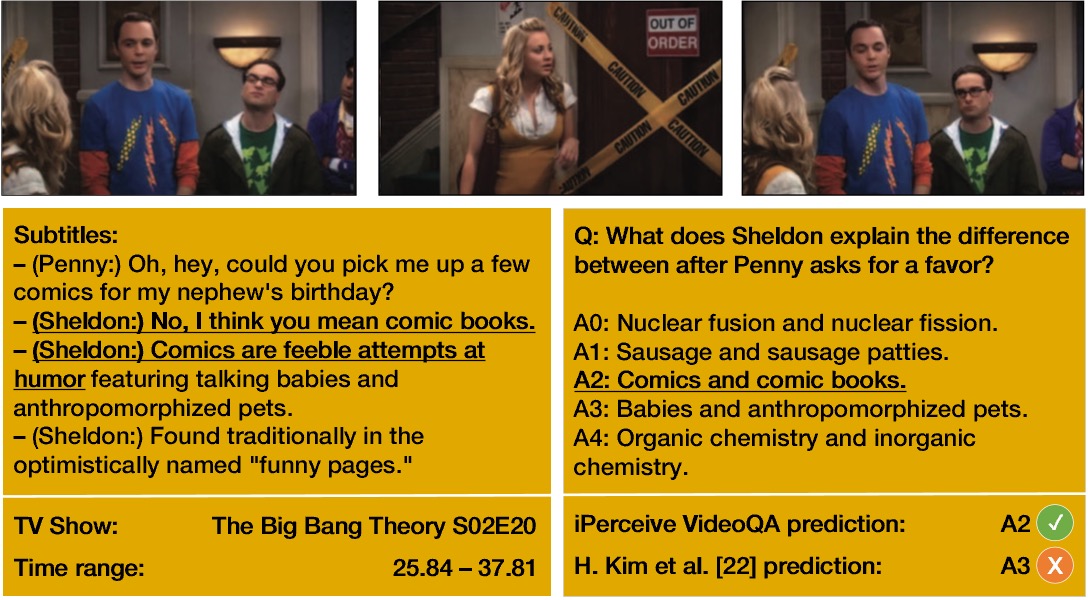} 
    \caption{Qualitative sampling of iPerceive VideoQA: Question-answering for a sample from the TVQA validation set.}
    \label{VidQAsample}
\end{figure}

Tab. \ref{VQAresults} shows a comparison of iPerceive VideoQA with the state-of-the-art. Fig. \ref{VidQAsample} offers a qualitative comparison between iPerceive VideoQA and \cite{kim2020dense} which was the best performing baseline at the time of writing. 

\begin{table}[!h]
\caption{Evaluation of iPerceive VideoQA on the TVQA dataset.}
\begin{adjustbox}{width=\columnwidth,center}
    \vspace {-2.5mm}
    \begin{tabular}
    {lcccccccc}
        \toprule
\multirow{2}{*}{Method} & \multicolumn{7}{c}{Test-Public (\%)} & \multirow{2}{*}{Val (\%)} \\
& Mean & BBT & Friends & HIMYM & Grey & House & Castle \\\hline\hline
Lei et al. \cite{lei2018tvqa} & 66.46 & 70.25 & 65.78 & 64.02 & 67.20 & 66.84 & 63.96 & 65.85 \\
J. Kim et al. \cite{kim2019progressive} & 66.77 & - & - & - & - & - & - & - \\
J. Kim et al. \cite{kim2019gaining} & 67.05 & - & - & - & - & - & - & - \\
J. Kim et al. \cite{kim2020modality} & 71.13 & - & - & - & - & - & - & - \\
Yang et al. \cite{yang2020bert} & 73.15 & - & - & - & - & - & - & - \\
H. Kim et al. \cite{kim2020dense} & 74.09 & 74.04 & 73.03 & 74.34 & 73.44 & 74.68 & 74.86 & 74.20
\\
iPerceive VideoQA & \textbf{75.15} & \textbf{75.32} & \textbf{74.22} & \textbf{75.14} & \textbf{74.42} & \textbf{75.22} & \textbf{75.77} & \textbf{76.97} \\
        \bottomrule
    \end{tabular}
\label{VQAresults}
\end{adjustbox}
\end{table}

\subsubsection{Ablation Analysis}
Tab. \ref{VQAab} shows ablation studies for iPerceive VideoQA to assess the impact of common-sense reasoning and iPerceive DVC as design decisions.

\begin{table}[!h]
\caption{Ablation analysis for iPerceive VideoQA.}
\begin{adjustbox}{width=0.7\columnwidth,center}
    \vspace {-2.5mm}
    \begin{tabular}
    {ccc}
        \toprule
Common-Sense Reasoning & iPerceive DVC & Val (\%) \\\hline\hline
\xmark & \xmark & 74.20\\
\xmark & \cmark & 75.42\\
\cmark & \xmark & 75.55\\
\cmark & \cmark & 76.97\\
        \bottomrule
    \end{tabular}
\label{VQAab}
\end{adjustbox}
\end{table}

\section{Conclusion}
We proposed iPerceive, a portable framework that enables common-sense learning for videos by building a knowledge base using contextual cues. We demonstrated the effectiveness of iPerceive on the tasks of DVC and VideoQA using the ActivityNet Captions and TVQA datasets, respectively. 
Furthermore, iPerceive DVC blends multi-modal DVC with end-to-end Transformer-based learning. 
Using ablation studies, we showed that these common-sense features help the model better perceive relationships between events in videos, leading to improved performance on challenging video tasks that need cognition. 

\section{Broader Impact}
Machines perceive their immediate world by analyzing videos of their environment, à la humans being cognizant of their surroundings. Video understanding is thus a critical area to conquer the perception gap between man and machine. Our work propels the idea of causal reasoning for machines and bring us a step closer to the ultimate goal of visual-linguistic causal reasoning, a distinct quality that makes us human. Since our work is easily portable, we hope that the promising results in our work would encourage researchers to further explore the idea of common-sense reasoning and apply it to new applications in the field of video and language understanding.

{\small
\bibliographystyle{ieee_fullname}
\bibliography{iPerceive}
}

\clearpage

\section*{8. Supplementary Material}
The supplementary material consists of two sections. In Section 8.1, we provide qualitative results of iPerceive DVC on another example video from the ActivityNet Captions validation set. In Section 8.2, we provide qualitative results of iPerceive VideoQA on additional samples from the TVQA test set. 

\subsection*{8.1 Qualitative sampling of iPerceive DVC}

\begin{figure*}[hb]
    \centering
    \includegraphics[width=1.00\linewidth]{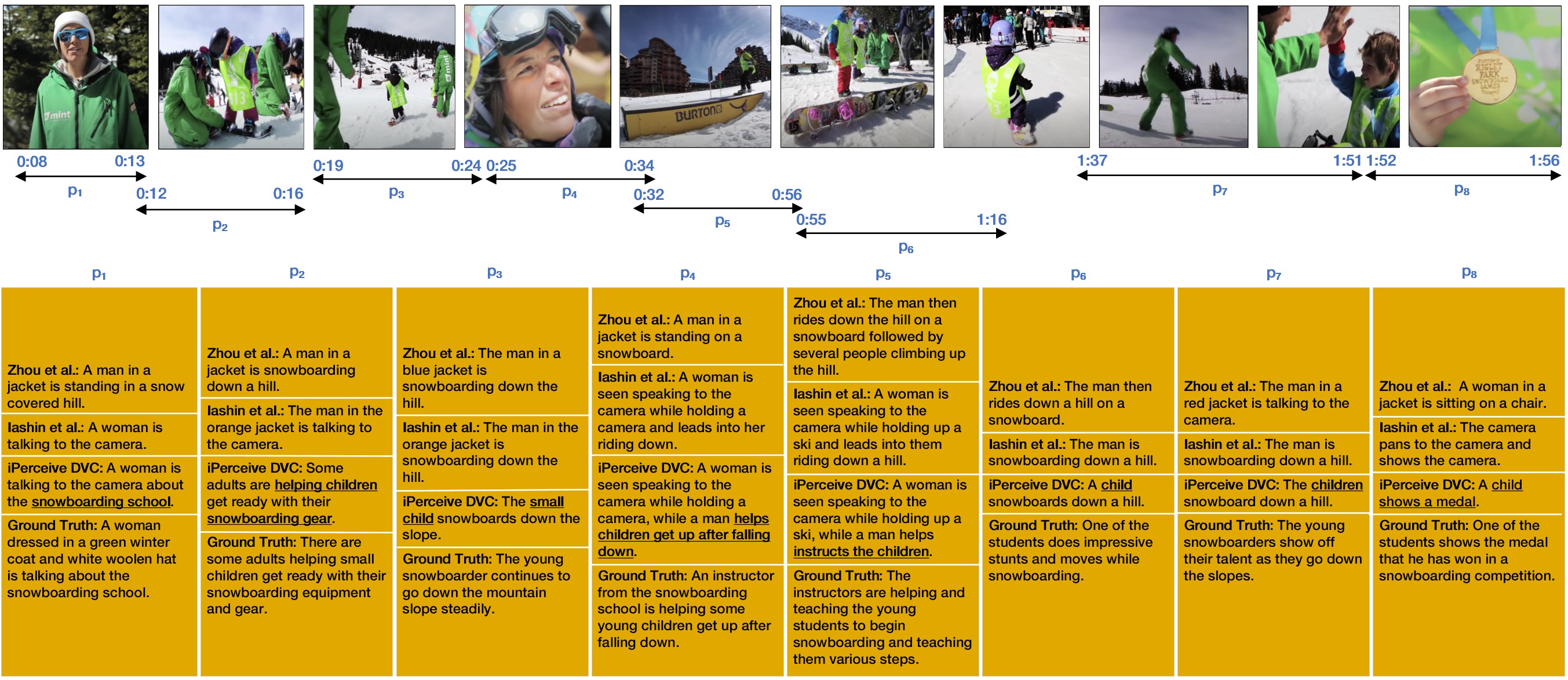} 
    \renewcommand{\thefigure}{6}    
    \caption{DVC results for a sample video from the ActivityNet Captions validation set with a multitude of overlapping and individual event proposals.}
    \label{DVCsample2}
\end{figure*}

Figure \ref{DVCsample2} shows a qualitative analysis of DVC using a sample from the ActivityNet Captions validation set which features $p_{1-8}$ as its event proposals. 

For each event proposal $p_n$, we compare the output of \cite{zhou2018end, iashin2020multi}, iPerceive DVC and the ground truth. The video (YouTube video id:  \texttt{EGrXaq213Oc}) lasts two minutes and contains 12 human annotations. The video is an advertisement for snowboarding lessons for children. It shows examples of children successfully riding a snowboard on a hill and supportive adults that help them to learn. A lady narrates the video and appears in the shot a couple of times.

From the captions of the event proposals, we can identify numerous instances where the fact that there exists causal relationships between events is helpful for the model. For instance, the fact that the children are being ``instructed'' because they are ``attending snowboarding school'' helps the model render a better output especially in event 5. A similar pattern can be observed in event 4 where ``children are getting up'' because they ``fell down''. Iashin et al. \cite{iashin2020multi} offer an in-depth treatment into the importance of additional modalities for dense video captioning, namely, speech and audio and how they enhance the articulation quotient of the output. 

\subsection*{8.2 Qualitative sampling of iPerceive VideoQA}

\begin{figure}[h]
    \centering
    \includegraphics[width=0.90\linewidth]{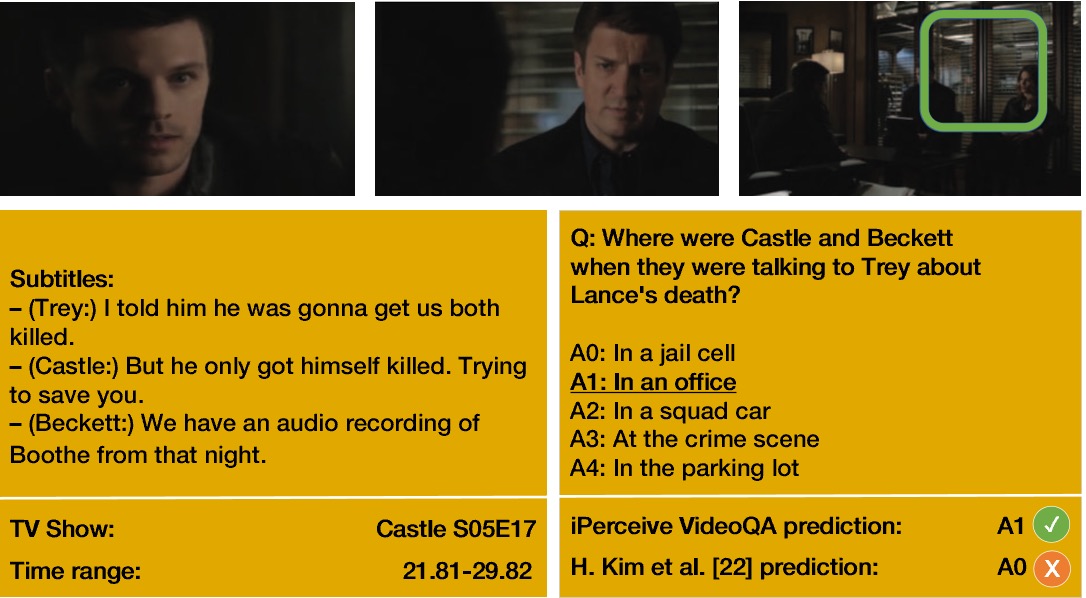} 
    \renewcommand{\thefigure}{7}    
    \caption{VideoQA for a sample from the TVQA test set featuring a Castle episode.}
    \label{VidQAsample3}
\end{figure}

\begin{figure}[h]
    \centering
    \includegraphics[width=0.90\linewidth]{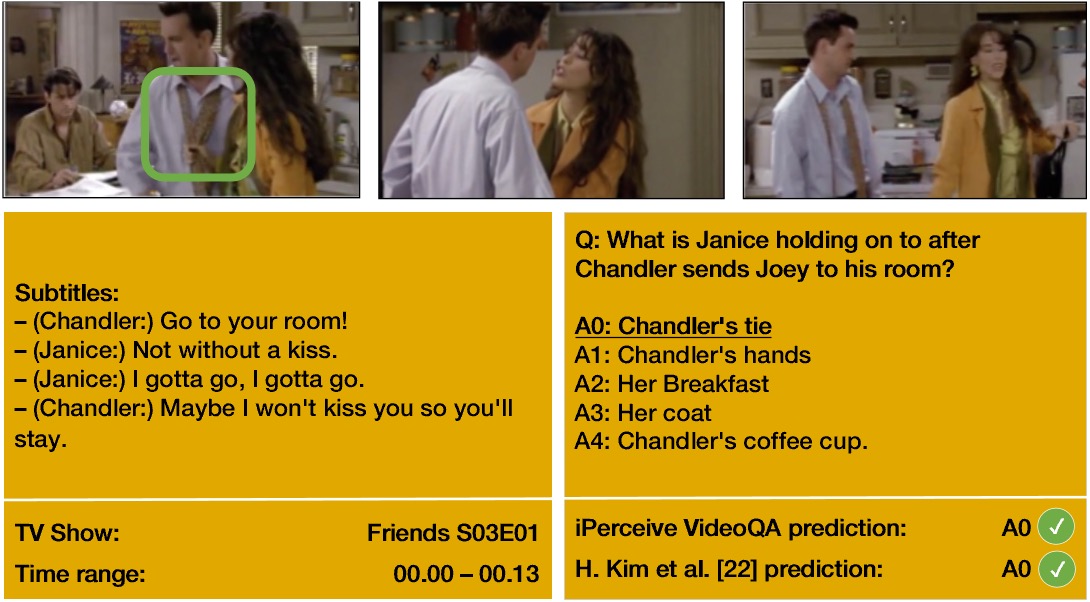} 
    \renewcommand{\thefigure}{8}    
    \caption{VideoQA for a sample from the TVQA test set featuring a Friends episode.}
    \label{VidQAsample2}
\end{figure}

\begin{figure}[h]
    \centering
    \includegraphics[width=0.90\linewidth]{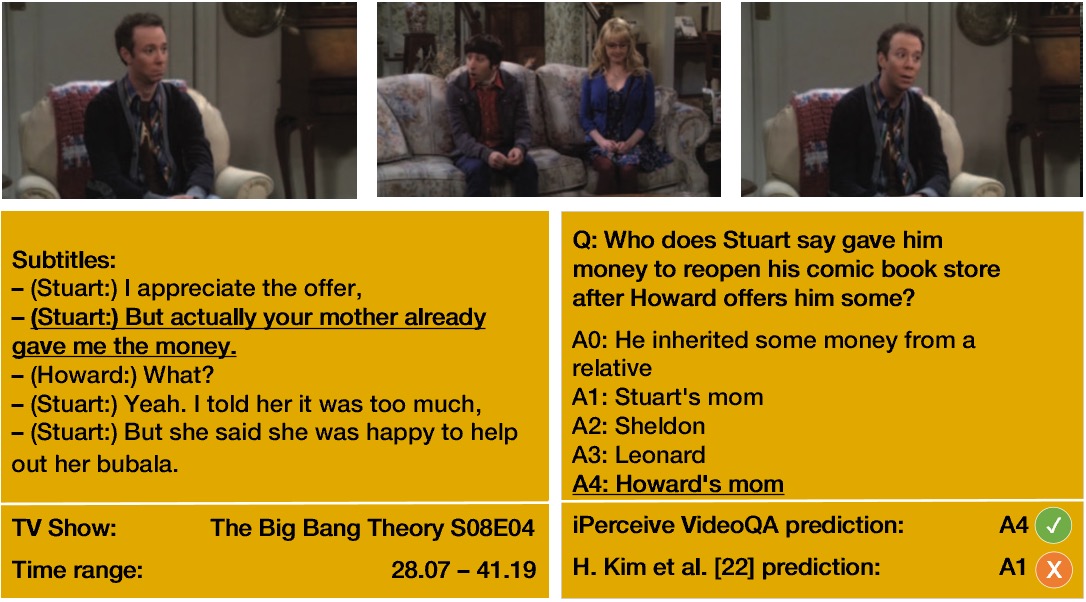} 
    \renewcommand{\thefigure}{9}    
    \caption{VideoQA for a sample from the TVQA test set featuring an episode from  The Big Bang Theory.}
    \label{VidQAsample5}
\end{figure}

\begin{figure}[h]
    \centering
    \includegraphics[width=0.90\linewidth]{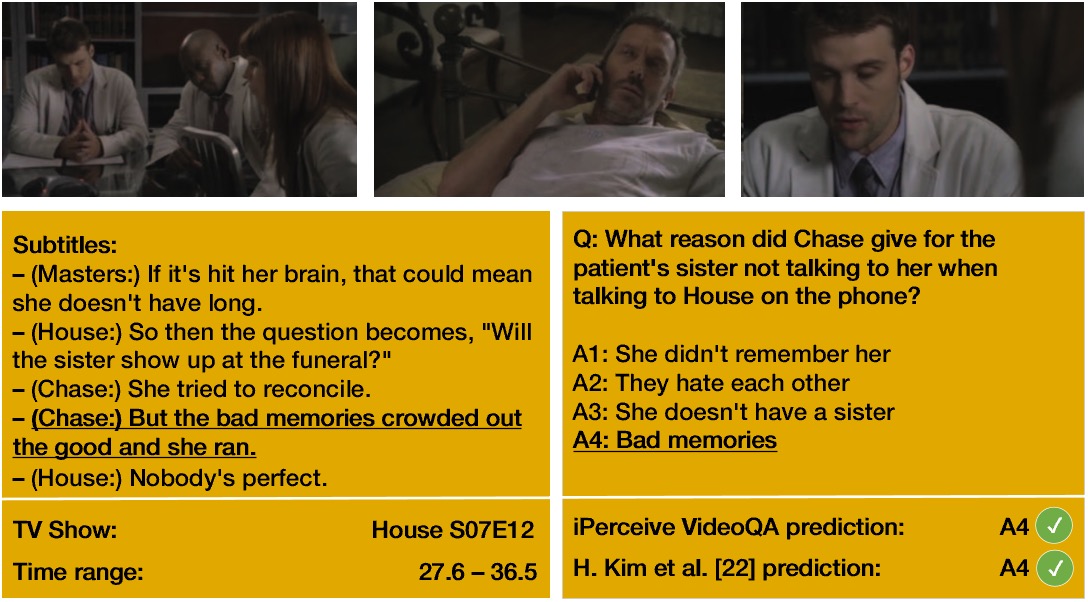} 
    \renewcommand{\thefigure}{10}    
    \caption{VideoQA for a sample from the TVQA test set featuring a House episode.}
    \label{VidQAsample4}
\end{figure}

Figures \ref{VidQAsample3}, \ref{VidQAsample2}, \ref{VidQAsample5} and \ref{VidQAsample4} show qualitative analyses of VideoQA using the TVQA test set featuring a Castle, Friends, The Big Bang Theory and House episode respectively.

For each sample, video frames with the corresponding subtitles and question-answers are included. The correct answers are highlighted for each sample. Note that the samples in Figures \ref{VidQAsample3} and \ref{VidQAsample2} show the attended visual area that the model uses to make its prediction. Similarly, the samples in Figures \ref{VidQAsample5} and \ref{VidQAsample4} highlight the subtitles that are relevant to the question at hand.

Similar to what we saw above with iPerceive DVC, common-sense reasoning helps correct the underlying visual attention while deriving causal relationships both within and between events. This manifests as improved accuracy for the VideoQA task. 

\end{document}